\makeatletter\def\graphicscache@inhibit{true}\makeatother
\documentclass{article}
\usepackage[utf8]{inputenc}
\usepackage{jfrExamplee}
\usepackage{graphicx}

\usepackage{natbib}
\let\bibhang\relax
\usepackage{apalike}

\usepackage{setspace}

\usepackage[hidelinks]{hyperref}
\usepackage[capitalise]{cleveref}

\usepackage{tikz,pgfplots,pgfplotstable}

\usepackage{amsmath}
\usepackage{textcomp}
\usepackage{subcaption}
\usepackage{threeparttable}

\usepackage{tikz}
\usetikzlibrary{arrows}
\usetikzlibrary{positioning,calc}
\usetikzlibrary{decorations.pathreplacing}
\usetikzlibrary{decorations.markings}
\usetikzlibrary{fit}
\usetikzlibrary{shapes.callouts}
\usetikzlibrary{shapes.geometric}
\usetikzlibrary{matrix}

\usepackage{pgfplots}
\pgfplotsset{compat=newest}
\usepgfplotslibrary{groupplots}
\usepgfplotslibrary{units}

\usepackage{booktabs}

\usepackage[tightpage]{preview}

\newenvironment{maybepreview}%
{\noindent\ignorespaces}%
{\par\noindent%
\ignorespacesafterend}
\PreviewEnvironment{maybepreview}

\usepackage[multidot]{grffile}

\usepackage{graphicscache}

\title{Team NimbRo at MBZIRC 2017:\\ Autonomous Valve Stem Turning using a Wrench}

\author{
Max Schwarz \\
Autonomous Intelligent Systems \\
University of Bonn \\
\texttt{schwarz@ais.uni-bonn.de} \\
\And
David Droeschel \\
Autonomous Intelligent Systems \\
University of Bonn \\
\And
Christian Lenz \\
Autonomous Intelligent Systems \\
University of Bonn \\
\And
Arul Selvam Periyasamy \\
Autonomous Intelligent Systems \\
University of Bonn \\
\And
En Yen Puang \\
University of Bonn \\
\And
Jan Razlaw \\
Autonomous Intelligent Systems \\
University of Bonn \\
\And
Diego Rodriguez \\
Autonomous Intelligent Systems \\
University of Bonn \\
\And
Sebastian Schüller \\
Technical Computer Science \\
University of Bonn \\
\And
Michael Schreiber \\
Autonomous Intelligent Systems \\
University of Bonn \\
\And
Sven Behnke \\
Autonomous Intelligent Systems \\
University of Bonn
}

\begin{document}

\maketitle

\begin{tikzpicture}[remember picture,overlay]
\node[anchor=north west,align=left,font=\sffamily,xshift=0.2cm,yshift=-0.2cm,text=gray] at (current page.north west) {%
  Accepted for Journal of Field Robotics (JFR), Wiley, to appear 2018.
};
\end{tikzpicture}%

\begin{abstract}

The Mohamed Bin Zayed International Robotics Challenge (MBZIRC) 2017 has defined ambitious new benchmarks to advance the state-of-the-art in autonomous operation of ground-based and flying robots.
In this article, we describe our winning entry to MBZIRC Challenge~2: the mobile manipulation robot Mario. It is capable of autonomously
solving a valve manipulation task using a wrench tool detected, grasped, and finally
employed to turn a valve stem.
Mario's omnidirectional base allows both fast locomotion and precise close
approach to the manipulation panel.
We describe an efficient detector for medium-sized objects in 3D laser scans
and apply it to detect the manipulation panel.
An object detection architecture based on deep neural networks is used
to find and select the correct tool from grayscale images.
Parametrized motion primitives are adapted online to percepts of the tool and
valve stem in order to turn the stem.
We report in detail on our winning performance at the challenge and discuss
lessons learned.

\end{abstract}

\section{Introduction}

The Mohamed Bin Zayed International Robotics Challenge (MBZIRC) 2017 presented
three challenging tasks for robotic systems. In addition to two challenges for
autonomous aerial vehicles (Challenge~1 and Challenge~3), Challenge~2 required a ground robot to locate a manipulation panel in a 60$\times$60\,m
square arena, to select an appropriate wrench fitting to a valve stem with square cross section, and finally
to operate the valve by turning it one full revolution with the wrench.
Such a task is relevant for many scenarios, such as maintenance and disaster
response in areas that are hard to reach or dangerous for humans.
In this way, the MBZIRC Challenge~2 extends the work started in recent disaster-response challenges like the DARPA Robotics Challenge \citep{pratt2013darpa}
and the ARGOS Challenge \citep{kydd2015autonomous}.
\section{Related Work}

Despite the short time between MBZIRC and the submission of this article, 
the first works reporting on MBZIRC systems have already been published.
\Citet{chen2016development}, for example, describe a combination of the versatile HRP-2
humanoid robot upper body with a wheeled mobile base specifically designed for the tasks of
the MBZIRC. While the base is capable of similar velocities as our robot of
up to $4\,\frac{\textrm{m}}{\textrm{s}}$, it uses skid steering and is thus non-omnidirectional
in contrast to our design.
For selecting the correct wrench, the team uses a Hough transform to
detect vertical lines and clusters them using k-means.
Note that this work was published before the MBZIRC competition and
may not reflect the final system. During the competition, team JSK
used teleoperation to complete the task, resulting in a fifth place
for Challenge~2.

Several robotic systems developed for the DARPA Robotics Challenge (DRC)
are able to perform similar tasks from a perceptual, kinematic,
and mechanical perspective.
In particular, one of the eight tasks required robots to
operate an industrial valve, although the valve featured a big
circular handle and thus did not require any tools.
However, a second DRC task required to use a drywall
cutting tool to cut out a circular hole.
The DRC did not require full autonomy, so the systems were
geared towards teleoperation.

\Citet{cho2011,kim2010} describe DRC-HUBO, the winning DRC system.
The humanoid robot is capable of walking and driving locomotion,
switching from walking to driving by kneeling down onto wheels
mounted at the knees and ankles.
Team IHMC~\citep{johnson2015team} came in second at the DRC finals.
Their system is fully humanoid and is only capable of walking.
The third place went to CHIMP~\citep{stentz2015chimp}, a roughly
humanoid robot capable of driving on all fours. Using belt drives
on its legs, CHIMP is able to turn in a skid-steering fashion.

Our own entry to the DRC, the Momaro system~\citep{schwarz2017nimbro}, achieved a
fourth place at the finals with seven out of eight tasks completed.
In contrast to the systems developed for the DRC, our robot Mario is capable
of fast locomotion and fully autonomous operation.

Our group investigated more autonomous behavior with Momaro in our
entry the DLR SpaceBot Camp 2015 \citep{Schwarz:Frontiers2016}.
Here, Momaro's versatile manipulation and locomotion capabilities
were extended with a control layer capable of autonomously executing complex missions
consisting of navigation, search, and manipulation actions.

The ARGOS challenge\footnote{\url{http://argos-challenge.com/}},
having taken place from 2013 to 2017, presented interesting
tasks in the context of industrial inspection. In contrast to the DRC,
it focused more on realistic scenarios (possibly volatile or
explosive oil \& gas facilities). The robotic systems
were only used for reconnaissance, however. Manipulation was not required. The winning team
Argonauts~\citep{kohlbrecher2016robocup, schillinger2016human}
utilized a tracked robot equipped with a multi-modal sensor arm.

More generally, robotic tool use is still a challenging problem.
Similar to many recent works \citep{hoffmann2014adaptive, stuckler2014adaptive},
our approach does not rely on high-precision grasping, but rather
is able to observe both the grasped tool with its tip and
the object to be manipulated in order to calculate
any correction needed due to imprecise grasping.
\pagebreak %
\section{Mechanical Design}

\begin{figure}
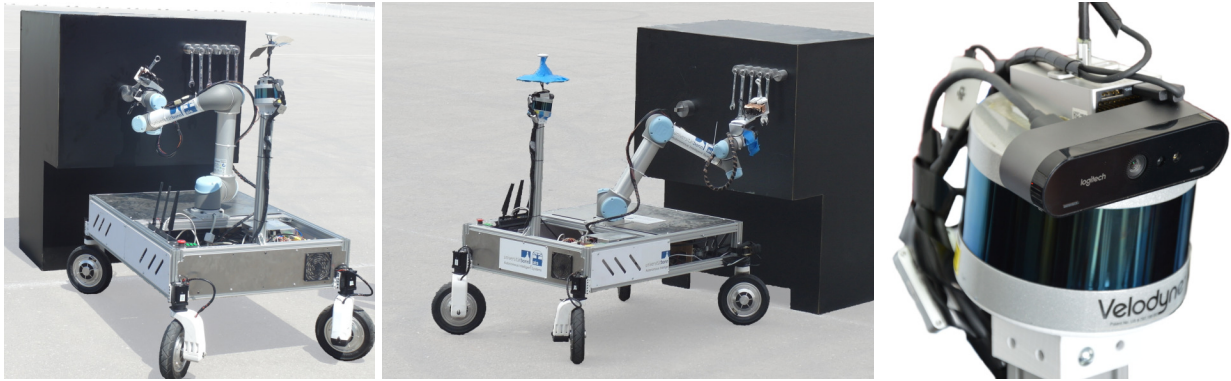

 \centering\begin{maybepreview}
 \includegraphics[height=5cm]{images/mario_faded_background.jpg}
 \includegraphics[height=5cm]{images/grasp_faded_background.jpg}
 \includegraphics[height=5cm]{images/sensor_head.jpg}\end{maybepreview}
 \caption{Left: Mario turning the valve during practice.
 Center: Mario grasping the correct wrench during the Challenge~2 finals.
 Right: Mario's sensor head with 3D laser scanner, DJI Flight Controller, and webcam.
 }
 \label{fig:mario_valve}
\end{figure}

Our robot Mario (see \cref{fig:mario_valve}) was specifically designed for
the requirements of MBZIRC Challenge~2.
In particular, we focused on two key design goals: speed and simplicity.
Since the task itself seemed manageable and had been demonstrated in different
variants (albeit under teleoperation) in earlier competitions such as the DRC,
we focused from the beginning on execution time as an important design
goal. On the other hand, keeping the robot design as simple as possible kept
the engineering overhead manageable and allowed us to make the system as robust
as possible in the short time frame before the competition.

\subsection{Omnidirectional Base}
\label{sec:base}

Since we expected mostly flat terrain, we opted for a wheeled base. To have a
stable support for manipulation and during fast driving, a four-wheeled configuration with 1.08$\times$0.78\,m footprint was chosen.
In our experience, an omnidirectional base offers two advantages in a mobile manipulation
scenario: It allows the robot to position its base very precisely and does not
need complicated maneuvering for small adjustments. For these reasons, we
incorporated omnidirectional bases into most of our previous mobile manipulation robots
\citep{Stueckler:Frontiers2016, schwarz2017nimbro}. An additional advantage in this
particular setting is that Mario is able to circle the valve panel while
facing it---allowing for very precise tracking of the panel during the approach.

As the distance that has to be traveled is quite large (the diagonal of the
arena measures 85\,m), a faster robotic base can yield significant savings in task
execution time. Our previous robots used servo motors with high gear ratios as drive
actuators, which was much too slow for this purpose. Instead, Mario is equipped
with direct-drive brushless DC hub motors inside each wheel. The motors were
originally intended for hover boards, i.e. personal conveyance devices, and thus
offer sufficient torque for accelerating the 65\,kg robot quickly as well as for reaching high maximum speeds.
We measured a stall torque of 10.3\,Nm per wheel without optimizing the wheel controllers
for this purpose. 
Using this base, Mario can exert 330\,N force or accelerate
with up to $5\,\frac{\textrm{m}}{\textrm{s²}}$ and reach velocities of up to $4\,\frac{\textrm{m}}{\textrm{s}}$.

\subsection{Sensor Head}

Mario carries a sensor head with sensors mainly used for navigation and
supervision (see \cref{fig:mario_valve}).
It consists of a Velodyne VLP-16\footnote{\url{http://velodynelidar.com/vlp-16.html}} 3D laser scanner coupled with
a DJI N3 Flight Controller\footnote{\url{https://www.dji.com/n3/info}}, which offers inertial, GPS, and compass measurements with a rate of 100\,Hz.

The VLP-16 is a multi-beam light detection and ranging (LiDAR) sensor that measures 300.000 distances (up to 100\,m) to surrounding objects per second. 
It has 16 laser/detector pairs placed on a vertical axis with a spacing of $2^\circ$, resulting in a $30^\circ$ vertical field of view.
This setup allows to get 16 measurements at a time. 
By spinning the laser/detector pairs around the vertical axis, a $360^\circ$ scan of the environment is achieved at a rate of, e.g., 10\,Hz. 

In addition to the distance measurements, the VLP-16 provides an intensity value for each point, according to the reflectivity of the surface that is measured.
The main advantages of this sensor are its high range of 100\,m, low weight, and low power consumption. 
Furthermore, it has a high accuracy of $\pm$3\,cm, a good initial calibration and is robust to temperature changes \citep{glennie2016calibration}.
The laser was mounted pitched downwards by $10^\circ$, which gives better measurement coverage in front of the robot.
Since our omnidirectional base allows us to keep the robot facing the panel during the approach, coverage behind the robot is not needed.

The sensor head also features a Logitech
webcam that is used for supervision of the robot, i.e. giving the operators
enough situational awareness to be able to stop the robot if something should go
wrong.

\subsection{Manipulator}

\begin{figure}
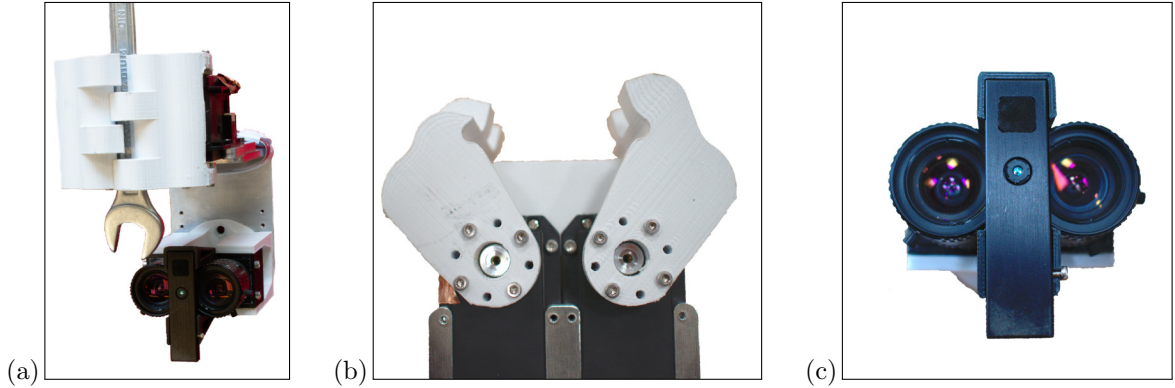

 \centering\begin{maybepreview}
 \setlength{\fboxsep}{0cm}
 (a)\,\fbox{\includegraphics[height=5cm]{images/gripper/20170808_0024_soft.jpg}}\hspace*{3ex}
 (b)\,\fbox{\includegraphics[clip,trim=200 0 100 0,height=5cm]{images/Gripper.jpg}}\hspace*{3ex}
 (c)\,\fbox{\includegraphics[clip,trim=200 0 200 0,height=5cm]{images/Kameras.jpg}}\end{maybepreview}
 \caption{Endeffector design. (a) Entire endeffector grasping a wrench;  
 (b) View of the gripper from below (note the interlocking fingers);
  (c) Camera module with two PointGrey grayscale cameras and the PMD CamBoard pico flexx
 camera in the center.}
 \label{fig:endeffector}
\end{figure}

Towards the simplicity design goal, we opted for an off-the-shelf robotic arm: Universal Robots UR5. The 6-DoF arm has sufficient working
range to grasp the wrench and turn the valve stem without a need for repositioning the robot
base. Furthermore, the arm offers full ROS integration and approximate contact
force/torque measurements.

We extended the arm with a custom endeffector (see \cref{fig:endeffector})
carrying a wrench gripper and
our object perception sensors: two Point Grey 2.3\,MPixel monocular cameras and a PMD CamBoard pico flexx
depth camera. The PointGrey cameras offer high-resolution stereo images, which
are used to detect and select the correct wrench
(see \cref{sec:wrench_perception}). Since the wrenches appear mostly
white-on-black, we opted for grayscale cameras to increase the available
resolution. The pico flexx camera is able to measure close-range depth even
under direct sunlight conditions, which allows us to perceive the valve stem
robustly.

The gripper is a custom designed 2-DoF pinch gripper, consisting of
two 3D-printed fingers. The fingers are specifically designed to grasp wrenches.
The gripper is positioned in such a way that the wrench mouth, when grasped,
is in the field of view of the pico flexx camera. This allows the system
to directly perceive the offset between the wrench mouth and the valve stem.

\subsection{Electrical System}

Mario is powered by an eight-cell LiPo battery with 16\,Ah and 29.6\,V nominal
voltage. The high voltage is necessary to drive the brushless wheel motors.
The battery allows for an hour of operation, depending on task intensity.
For continuous operation, the battery can be hot-swapped without shutting
down the system.

The UR5 arm includes a heavy control box and operator panel, and requires
a 230\,V supply. We removed the outer
industrial-strength enclosure of the control box and replaced an internal
voltage converter to be able to run the system directly from the battery.
The operator panel was also removed to save weight.

The robot is controlled by an onboard computer consisting of a standard ATX
mainboard equipped with a quad-core Intel Core i7-6700k CPU and 64\,GB RAM.
For inference with deep neural networks, a compact NVidia GeForce GTX 1050
Ti card with 4\,GB RAM is used.

The system can be remotely monitored using a WiFi connection. For this purpose,
it carries a Netgear Nighthawk AC1900 router, which is operated in client
configuration.
\section{Navigation and Panel Detection}
\label{sec:navigation}

The approach to the panel can be divided into three phases: i) allocentric
navigation, ii) navigation towards a panel detection, and iii) close approach of a
registered panel pose.

\subsection{Laser- and GPS-based Allocentric Navigation}

During the competition, it was revealed that the panel would be placed
along an arc with approximately constant distance to the starting location.
In order to find it, Mario drives along waypoints defined on an allocentric frame
while continuously performing panel detection (see \cref{sec:long_range_detection}).
The wheel odometry, which may be prone to drift, is corrected on every receipt of GPS
and compass measurements. Since GPS and compass are already filtered by the DJI N3 Flight Controller,
we perform no additional filtering.
To compensate for GPS uncertainty and drift, we employed a second GPS receiver
placed near the starting point at a known location, transmitted its GPS
pose over WiFi to Mario, and used the relative offset as the localization pose.
Note that the LiDAR is not used for localization, since the GPS precision suffices for
following predefined search waypoints until the panel is detected.
Similarly, a map of the arena (beyond the predefined waypoints) was not required for navigation.

Navigation is carried out using standard ROS planners for global and local
navigation, using Dijkstra and trajectory rollout techniques, respectively.
The cost map needed for navigation planning is dynamically generated from obstacles detected
in the laser scans, similar to the method described in \cref{sec:long_range_detection}.
Note that the final challenge terrain unexpectedly did not contain any obstacles,
so we turned most obstacle avoidance off for increased speed in the competition.

\subsection{Long-range Panel Detection}
\label{sec:long_range_detection}

\begin{figure}
\centering\begin{maybepreview}%
  \includegraphics[width=0.8\linewidth]{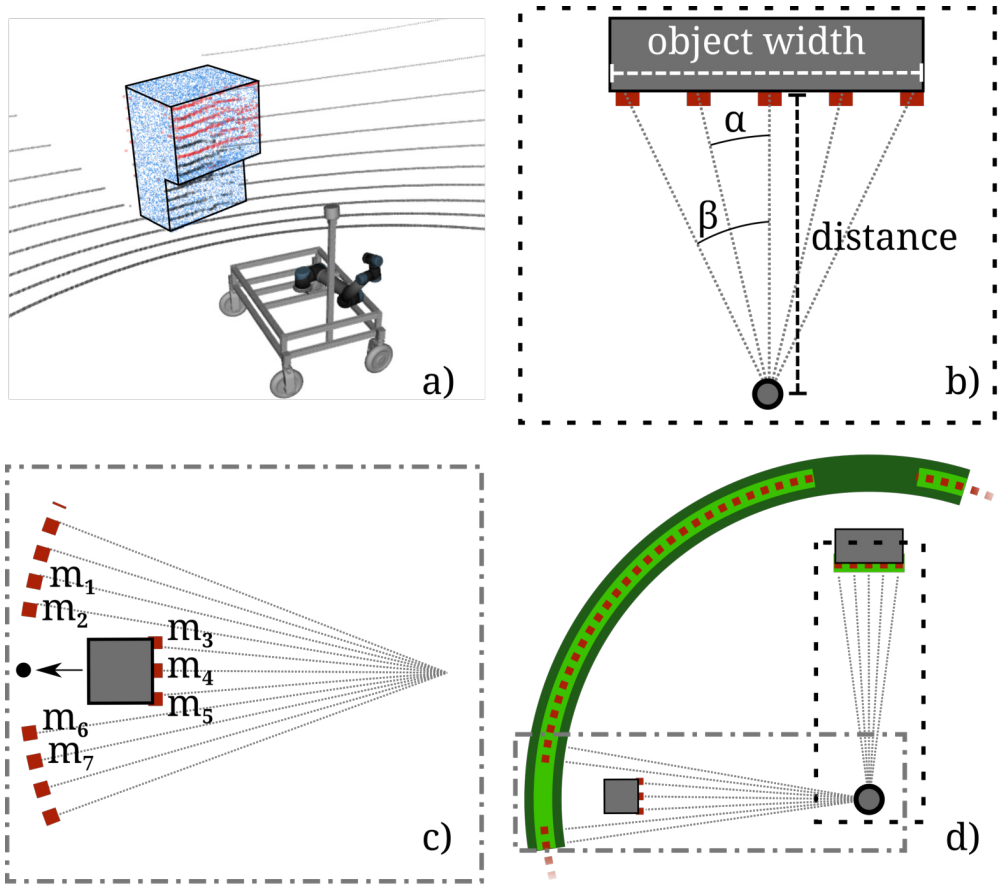}\end{maybepreview}
\caption{Panel detection.
\textbf{a)} Velodyne mounted on top of Mario in front of the panel (blue). Scan rings (black) get deformed by panel resulting in detections shown in red.
\textbf{b)} Conversion of object width to kernel size. The specified object width $w$ and the distance $d$ to the measurement are utilized to compute $\beta = \arctan(\frac{0.5w}{d})$. By using the fixed angle $\alpha$ between the measurements, we are able to approximate the upper threshold of the noise kernel size.
\textbf{c)} Example of a median filter applied on the middle point $m_4$ using a noise kernel size that exceeds the threshold. Three of the four background points $m_{1},m_{2},m_{6},m_{7}$ cancel out with the closer object points $m_{3},m_{4},m_{5}$ resulting in a filtered distance that is equal to one of the background points for $m_4$ (black dot).
\textbf{d)} Schematic example of the results of the median filters on a scan ring. Measurements are depicted as red squares. The result of the background kernel is represented by the dark green curve. The result of the noise kernel shown in bright green. The small object on the left is filtered by both medians, while the desired object is only filtered by the background kernel, forming a detection as the only difference between the median results.}
\label{fig:kernel_size}
\end{figure}

For detecting the panel, we process 3D scans from the Velodyne LiDAR sensor. We group measurements by the laser/detector pair they are originated from. 
Measurements from the same laser/detector pair for one full revolution are coined a \textit{scan ring}.
Objects in the scene deform the scan rings, resulting in measurements that are closer to the sensor than neighboring measurements from the background (Figure \ref{fig:kernel_size}a).
We aim for detecting groups of measurements that correspond to objects of a specified width. 

Our object detection approach applies a combination of median filters to the distance measurements of each scan ring individually. 
For detecting objects with a specific width, we apply a median filter with two different kernel sizes. 
A smaller kernel (\textit{noise kernel}) is used to remove noise and measurements belonging to smaller objects, while the larger kernel (\textit{background kernel}) captures the background. 
Objects of a specific size stick out of the background and result in response differences between the two filter kernels. 

We approximate the kernel size of the median filter based on the size of objects we want to detect, the measured distance, and the angle between neighboring measurements of the sensor (Figure \ref{fig:kernel_size}b). 
While the object size and the angle between neighboring measurements are constant, the distance to the object varies between measurements.
In other words, the kernel sizes are different for individual measurements in the ring.
To assure computation in real time, we apply a heuristic with simplified geometric assumptions, for instance by neglecting the object orientation.
The heuristic approximates the number of measurements reflected by the object. 
From the number of expected measurements from the heuristic, we directly deduce the noise kernel size by doubling this number (Figure \ref{fig:kernel_size}c) and approximate the background kernel by multiplying that size with a factor of $1.5$.

For every point in the scan ring, we calculate the response of the median filter with the noise kernel and the background kernel. 
For background points as well as points on smaller and larger objects, the filter response of both kernels is nearly equal. 
In contrast, objects with the desired size will result in a difference between the two filter responses (Figure \ref{fig:kernel_size}d). 
Since the detected object can be assumed to be closer to the sensor than the background, we filter out measurements where the response of the noise kernel is farther away than the response from the background kernel.
Measurements where the difference in filter responses exceeds a threshold are assumed to belong to an object.
This threshold was determined by experiments as $\kappa=0.61\,\textrm{m}$.

After processing each ring individually, we use Euclidean clustering to aggregate measurements from the different scan rings by their spatial distance. 
Furthermore, we apply a set of simple geometric filters on the resulting clusters, neglecting clusters that exceed the dimensions of our panel model. 
The resulting clusters are geo-referenced by means of GPS information and tracked over multiple frames. This allows us to neglect clusters outside the arena.

\subsection{Panel Registration}

\begin{figure}
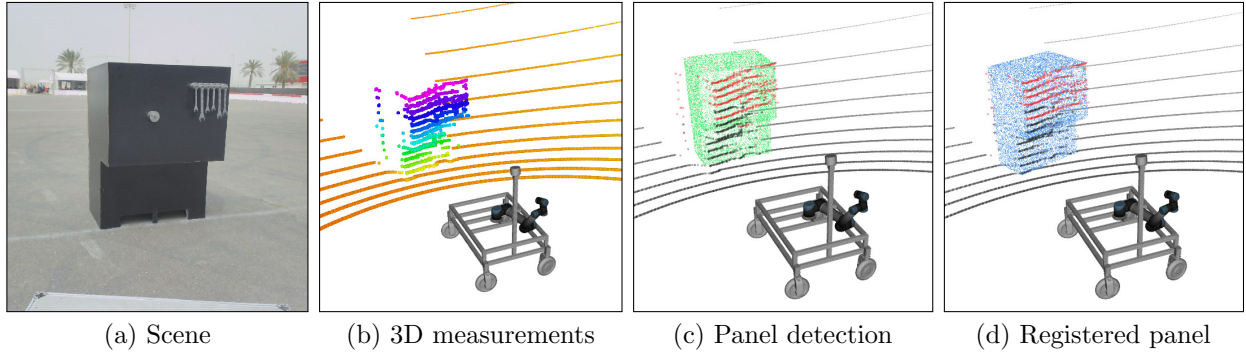

 \centering\begin{maybepreview}%
 \setlength{\fboxsep}{0cm}%
 \setlength{\tabcolsep}{2pt}%
 \begin{tabular}{cccc}%
  \fbox{\includegraphics[clip,trim=150 0 100 0,height=4.1cm]{images/panel_registration/panel_rgb.png}} &
  \fbox{\includegraphics[clip,trim=350 150 400 0,height=4.1cm]{images/panel_registration/panel_cloud.png}} &
  \fbox{\includegraphics[clip,trim=350 150 400 0,height=4.1cm]{images/panel_registration/panel_initial_guess.png}} &
  \fbox{\includegraphics[clip,trim=350 150 400 0,height=4.1cm]{images/panel_registration/panel_icp_result.png}} \\
  (a) Scene & (b) 3D measurements & (c) Panel detection & (d) Registered panel\\
 \end{tabular}\end{maybepreview}
 \caption{Panel registration pipeline: (a) webcam RGB image; (b) perceived pointcloud from the VLP-16 laser scanner (color depicts height); (c) initial panel model (green point cloud) and the panel detection output (red points); (d) ICP result with registered panel model.
 }
 \label{fig:panel_registration}
\end{figure}

After arriving at the panel, Mario has to face its front side for the manipulation subtask,
which means that we have to estimate the panel yaw angle.
We reuse the computed object clusters and align them to a panel model represented as a point cloud using the Iterative Closest Point (ICP) algorithm~\citep[see \cref{fig:panel_registration}]{holz2015registration}.

The cluster closest to the robot with a mean position above the ground plane is selected for registration. We rotate the panel model cloud successively around the vertical axis by $30^\circ$ and use 
each pose as an initial guess for ICP. The resulting transformation is restricted to translation on the ground plane and rotation around the vertical axis,
since the panel was always standing on the ground with the front side facing an unknown horizontal  direction.

Next, the different ICP results are rated with respect to a combined score. The first partial score is the weighted sum of mean distances between a panel model point and its
nearest cluster point. Since considering panel points from the invisible back face results in large errors, only points visible from the sensor head are used. To generate this
visibility information, a range image in the laser scanner frame is computed.

For the second partial score, we use the weighted sum over all distances now for each cluster point to its nearest panel model point.
Since the panel is largely symmetrical and it can be hard to distinguish between the front and back sides, we put higher weighting on
points close to the ground. This puts strong emphasis on the lower parts of the panel, where the largest difference in shape
between front and back side is located (see \cref{fig:panel_registration}).
Both scores added together result in the combined score for choosing the current panel pose. 
Finally, a low pass filter on the panel pose parametrized 
with Mario's angular velocity is applied to filter out large jumps in the panel orientation.

\begin{figure}
 \centering\begin{maybepreview}%
\begin{tikzpicture}[
      font=\footnotesize,
      trajlabel/.style={font=\scriptsize},
]

\begin{axis}[
	x=0.7cm, y=0.7cm,
    xmin=-10,
    height=6cm,
    width=\linewidth,
    colormap/jet,
    colorbar,
    traj/.style={mesh,point meta=explicit,line width=1pt},
    xlabel={X distance [m]},
    ylabel={Y distance [m]},
    colorbar style={ylabel={Velocity [m/s]}},
    point meta max=2
]

\addplot[traj] table[x=rx,y=ry,meta=velocity] {images/approach/run2_2017-03-17-10-45-54_001.txt};

\addplot[magenta,thick,opacity=0.6,quiver={u=\thisrow{ru},v=\thisrow{rv},scale arrows=0.6},-latex] table[x=rx,y=ry] {images/approach/run2_2017-03-17-10-45-54_001_sparse.txt};

\begin{scope}[shift={(axis direction cs:-1.1,0)}]

\fill[black] (axis cs:-0.38,-0.5)  rectangle (axis cs:0.38,0.5);

\end{scope}

\end{axis}

\end{tikzpicture} %
\end{maybepreview}
 \caption{Odometry record of the panel approach trajectory during our second
  competition run. The robot orientation is shown with magenta arrows pointing
  forward, spaced 1\,m apart along the trajectory. The panel is depicted as a black box.}
 \label{fig:approach_trajectory}
\end{figure}
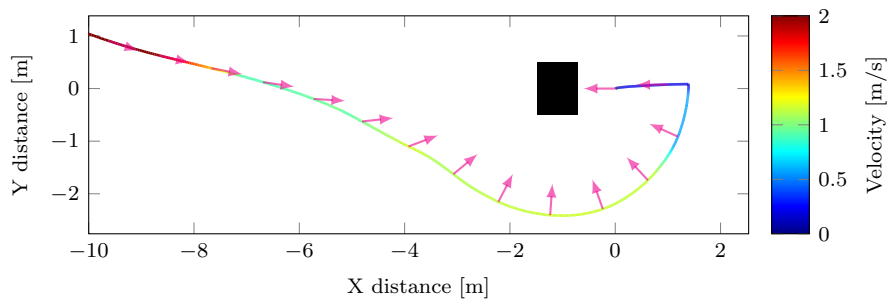

During approach to the panel, the panel pose is continuously transformed into an egocentric frame using odometry information to ensure a smooth approach.
The approach trajectory makes full usage of Mario's omnidirectional base.
The robot is always kept facing the panel using a simple proportional controller
on the yaw velocity. Linear velocity $(v_x,v_y)$ is controlled as follows:
\begin{alignat}{2}
 v_x &= \min \{ v_{\textrm{limit}}, p_x - d \} \\
 v_y &= \alpha \textrm{ rsgn} ( p_\theta ) d \frac{2 \pi}{T},
\end{alignat}
where the $x$-axis points forward, $v_{\textrm{limit}}$ denotes a velocity limit
of $0.8\,\frac{m}{s}$, $p$ is the detected panel position, $p_\theta$ the
panel yaw angle relative to the robot, $d=2.5\,\frac{m}{s}$ the desired circling radius,
and $T=10\,s$ denotes the duration of one turn around the panel.
We use a robust signum function $\textrm{rsgn}$ with an inbuilt $20^\circ$ hysteresis to
prevent oscillations when the panel is approached from its back.
Finally, $\alpha$ is used to smoothly enter the circling motion by fading it
from 0 to 1.

As soon as the angle $p_\theta$ is sufficiently small, the controller switches
to local approach to a pose directly in front of the panel. For this, the
3D $(x,y,\theta)$ target pose in egocentric coordinates is directly applied
as local velocity $(v_x, v_y, v_\theta)$. The approach is ended when the pose
difference is small enough. \Cref{fig:approach_trajectory} shows the resulting
approach trajectory.
In our opinion, the simplicity of this controller highlights the advantages of our
omnidirectional drive in situations requiring fast and precise object approach.
\section{Wrench and Valve Perception}
\label{sec:wrench_perception}

The challenge poses two object perception tasks: First, the correct wrench has
to be selected and localized among the six wrenches hanging on the panel.
Second, the wrench has to be correctly inserted onto the valve stem, which
requires precise estimation of the wrench mouth and valve stem poses.

\subsection{Wrench Ending Detection}

\begin{figure}
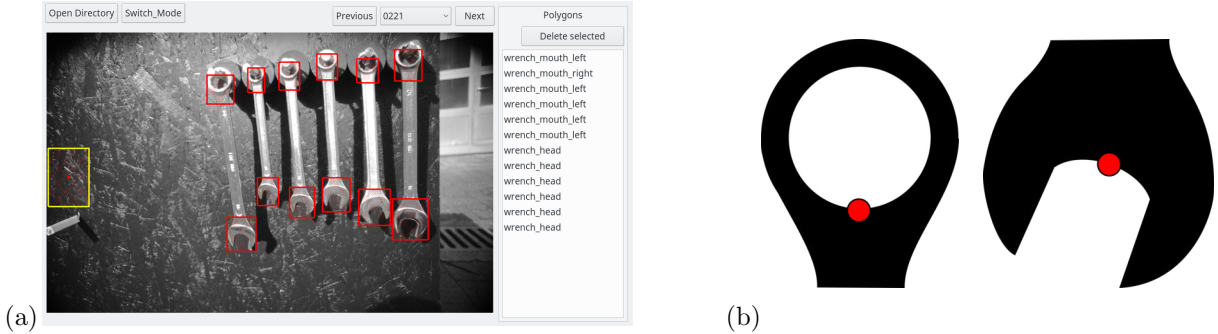

 \centering\begin{maybepreview}
 (a)\,\includegraphics[height=4.3cm]{images/annotation_tool.png}
 \hspace{1cm}
 (b)\raisebox{0.5cm}{\includegraphics[height=3.3cm]{images/annotation_head.png}}
 \hspace{3pt}
 \raisebox{0.5cm}{\includegraphics[height=3.3cm]{images/annotation_mouth_flop.png}}\end{maybepreview}
 \caption{Manual annotation of the wrench ends.
 (a) Annotation GUI; (b) Definition of bounding box centers
 on wrench head and mouth ends.
 }
 \label{fig:annotation}
\end{figure}

For the first subtask, i.e. selecting the correct wrench, we decided to use
a deep learning pipeline for object detection. We adapted our previous work
described by \citet{schwarz2016rgb}, which builds upon the DenseCap pipeline
developed by \citet{johnson2016densecap}. Briefly explained, the pipeline
follows the Faster R-CNN \citep{girshick2015fast} regime by creating rectangular
region proposals using a proposal network, then using bilinear interpolating
intermediate feature representations to normalize the proposal regions, and finally
classifying each region independently. For details, we refer to
\citet{schwarz2016rgb}.

We train the pipeline to detect two classes, namely the wrench ring piece end
(called ``head'' in the following) and the open end (``mouth''). We use the
pretrained DenseCap pipeline, which includes statistics learned from
ImageNet and the Visual Genome Dataset~\citep{krishna2017visual}, and finetune
on a set of 100 manually annotated images (see \cref{fig:annotation}) captured
using Mario's endeffector cameras.

The network predicts bounding boxes with parameters $(x_c,y_c,w,h)$ describing
the center point and size. The center point is defined consistently on the
wrench silhouette (see \cref{fig:annotation}). The bounding box size is annotated only
approximately to fit the head or mouth size, since it is not used later on.

\subsection{Wrench Size Estimation \& Selection}

\begin{figure}
 \centering\begin{maybepreview}
 \includegraphics[width=5cm]{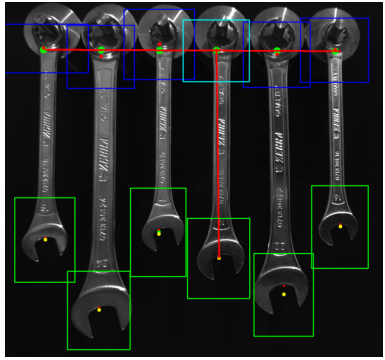}\end{maybepreview}
 \caption{Detected wrenches. Wrench heads are marked in blue, mouths in green.
 The regressed line through all head detections is shown in red. The correct
 wrench to grasp is also marked with a vertical red line.
 }
 \label{fig:wrench_matching}
\end{figure}

After detecting the head and mouth ends of the wrenches, the detections are
postprocessed: Since the wrenches are all mounted at the same height,
the head detections are refined using line regression. We allow the mouth
detections to move a small distance in order to ``snap'' to the nearest
vertical gradient maximum (i.e. the nearest horizontal edge).

Head and mouth detections are then greedily matched to form wrench pairs. The
matching cost is the horizontal deviation, i.e. we assume that the wrenches
are mostly hanging down with gravity. After matching head and mouth pieces and
thus retrieving a list of wrenches, we match the observed wrench lengths
to the expected lengths. The necessary depth information is obtained from
the ICP registration of the panel, under the assumption that the wrenches are
close to the panel surface. The correct wrench is then selected using the
closest match (see \cref{fig:wrench_matching}).

We repeat the entire wrench detection and selection process for both PointGrey
cameras and thus obtain two wrench poses. Only if both agree, i.e. the 3D wrench
head positions agree with a maximum tolerance of 2.5\,cm, the mean pose is taken
and used for grasping. Otherwise, the pipeline is run on new images. If this does not succeed within few seconds, the panel approach is triggered again to find a new camera pose and
hopefully improve the situation.

\subsection{Valve Stem Pose Estimation}
\label{sec:valve_perception}

\begin{figure}[b]
 \centering\begin{maybepreview}
 \includegraphics[height=5cm]{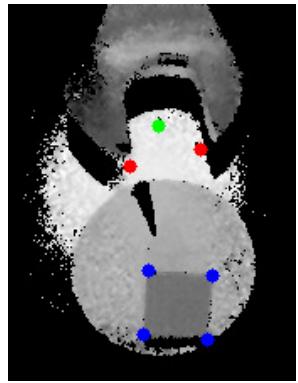}\end{maybepreview}
 \caption{Valve stem pose estimation. The detected wrench tips are shown as red
 points. The green point denotes the axis around which the wrench would turn.
 The minimum-size rotated bounding box around the valve stem front face is shown
 with blue dots at the four corners.
 }
 \label{fig:valve_stem}
\end{figure}

The second important perception task is to gather enough information to
precisely insert the wrench onto the rectangular valve stem. For this purpose,
a depth image captured by the pico flexx endeffector camera is used
(see \cref{fig:valve_stem}).
As a first step, a simple thresholding operation extracts the foreground
pixels---belonging either to the valve stem or the wrench mouth.
The wrench mouth is extracted using Euclidean clustering, selecting the
biggest cluster in the upper half of the image (see \cref{fig:valve_stem}).
Note that the wrench is fixed w.r.t. the camera. The two wrench tips are
extracted as local minima in $y$ coordinates.

The valve stem front face is then extracted as the largest cluster in the
lower half of the image. A 2D rotating caliper algorithm \citep{preparata1985computational} is used to fit a
minimum size rotated bounding box around the valve stem points, projected onto
the camera plane. The orientation of the valve stem and the 3D position of the
front face center point is then reported as a single 6D pose.
An exemplary output of the wrench and valve perception
can be seen in \cref{fig:valve_stem}.
\section{Manipulation}

The challenge mandated two main manipulation tasks: Grasping the selected wrench and turning the valve with it.
More precisely, the arm unfolds, is brought into a position to observe the wrenches,
grasps the selected wrench, inspects the valve stem with the endeffector sensors, aligns with the valve stem axis,
inserts the wrench, and executes a turning motion.
For all manipulation actions, we used parametrized motion primitives, which will be described in the following.
The insertion of the wrench onto the valve stem required particular attention and is detailed thereafter.

\subsection{Parametrized Motion Primitives}

The UR5 arm and the pinch gripper are controlled with parametrized motion primitives. 
This motion generation and execution pipeline was used in other robotic systems in our group before (see \cite{schwarz2017nimbro} and \cite{Schwarz:Frontiers2016}).
Motions are defined by a sequence of keyframes and are either generated at runtime or manually designed beforehand and
partially adapted to specified perception results before execution.

\begin{figure}
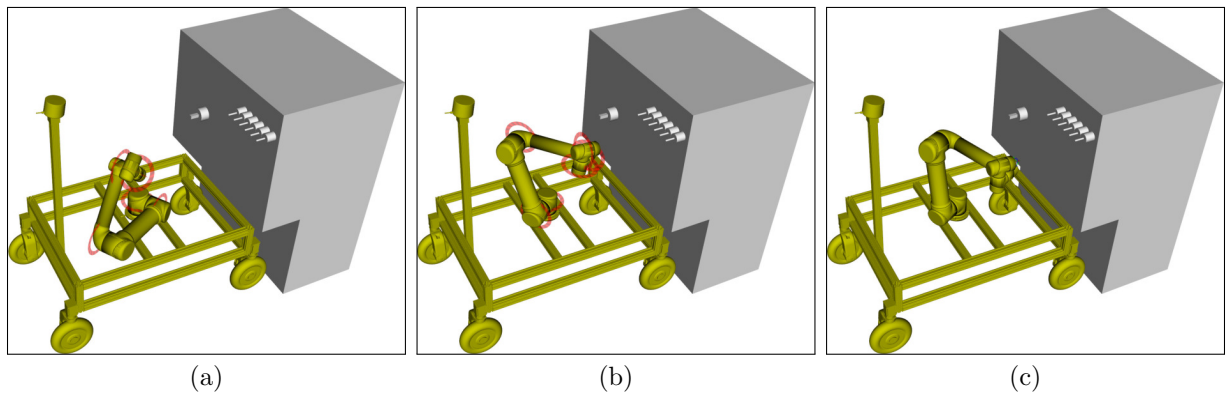

 \centering\begin{maybepreview}%
 \setlength{\fboxsep}{0cm}%
 \setlength{\tabcolsep}{2pt}%
 \begin{tabular}{ccc}
  \fbox{\includegraphics[clip,height=4.6cm]{images/keyframes/motion_1.jpg}} &
  \fbox{\includegraphics[clip,height=4.6cm]{images/keyframes/motion_2.jpg}} &
  \fbox{\includegraphics[clip,height=4.6cm]{images/keyframes/motion_3.jpg}} \\
  (a) & (b) & (c)\\
 \end{tabular}\end{maybepreview}
 \caption{Keyframes for extending the arm to perceive the wrenches with the Point Grey cameras: (a) and (b) moving the arm in joint-space; 
 (c) positioning the arm relative to the detected panel in Cartesian space.
 }
 \label{fig:keyframes}
\end{figure}

Mario's manipulator is separated in two kinematic groups: arm and gripper. 
Each keyframe consists of an arbitrary set of kinematic joint groups with an interpolation space (Cartesian or joint space). The interpolation between the 
current and desired joint-space position generates a trajectory which is checked for self-collisions and then sent to the hardware controller for execution.
For keyframes with Cartesian interpolation, the interpolated endeffector positions are mapped to joint configurations using an inverse kinematics solver.
In some cases, the Cartesian goal pose of a keyframe is
modified with perceived sensor data before interpolation. For example, for perceiving the wrenches, we want the endeffector in a fixed pose relative to the panel.
The corresponding keyframe is defined w.r.t. a reference panel pose. On perceiving the real panel pose, the keyframe is then shifted accordingly.
\Cref{fig:keyframes} shows the motion for moving the arm into the capture pose in front of the wrenches.

\subsection{Valve Stem Turning}

The last part of the MBZIRC Challenge~2 was to turn the valve stem clockwise by $360^\circ$ with the grasped wrench. 
We decided to use the wrench's mouth for operating the valve stem because it allows for insertion from the side, which seemed to require less accuracy than inserting the ring end of the wrench from the front.

Our first approach was to insert the wrench exactly onto the quadratic valve stem using force control provided by the UR5 arm and all necessary 
parameters, i.e. valve stem pose and wrench end points generated by the perception. This method did not work well, since the captured data and the force control mode 
were not as precise as expected.

\begin{figure}
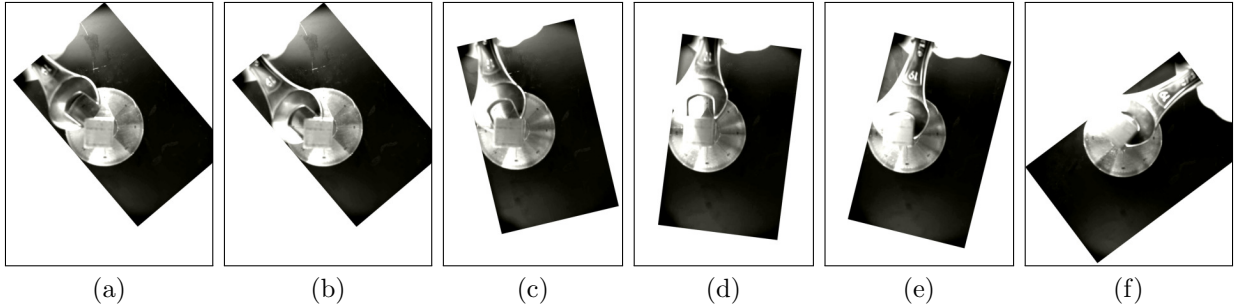

 \centering\begin{maybepreview}%
 \setlength{\fboxsep}{0cm}%
 \setlength{\tabcolsep}{2pt}%
 \begin{tabular}{cccccc}
  \fbox{\includegraphics[clip,trim=300 100 200 100,height=3.49cm]{images/valve/frame_0740_bright.png}} &
  \fbox{\includegraphics[clip,trim=300 100 200 100,height=3.49cm]{images/valve/frame_0745_bright.png}} &
  \fbox{\includegraphics[clip,trim=450 100 200 100,height=3.49cm]{images/valve/frame_0758_bright.png}} &
  \fbox{\includegraphics[clip,trim=450 100 200 100,height=3.49cm]{images/valve/frame_0767_bright.png}} &
  \fbox{\includegraphics[clip,trim=400 100 200 100,height=3.49cm]{images/valve/frame_0770_bright.png}} &
  \fbox{\includegraphics[clip,trim=300 100 200 100,height=3.49cm]{images/valve/frame_0787_bright.png}} \\%
  (a) & (b) & (c) & (d) & (e) & (f) \\%
 \end{tabular}\end{maybepreview}
 \caption{Inserting the wrench onto the valve stem.
  The images were captured by one of the Point Grey endeffector cameras and are
  rotated such that gravity points downwards.
  (a) close approach,
  (b) gripper is opened, wrench rests on stem, (c-d) turning with open gripper,
  (e) wrench inserted, (f) turning with closed gripper. 
 }
 \label{fig:inserting_wrench}
\end{figure}

To mitigate this issue, we developed a wrench insertion method that makes use of gravity (see \cref{fig:inserting_wrench}). First, we use the perceived valve stem orientation to approach it such that the wrench is rotated by $45^\circ$ from the
perfect insertion orientation. Next, we slightly open the gripper such that the wrench can now slide---driven by gravity---onto the valve stem.
The last step is to slowly rotate the wrench in this loose configuration around the valve stem. Gravity pushes the leading tip of the mouth onto the valve stem edge while the inner edge of the trailing mouth tip self-aligns with the valve stem corner which is inside the mouth. This process continues until the leading mouth tip passes the opposite valve stem corner and the entire wrench falls onto the valve stem. 
We maximize the chance of a successful insertion by covering the whole upper $180^\circ$ range, i.e. turning to one side until the wrench is
horizontal and then to the other side until the wrench is horizontal. Depending on the valve stem orientation, this yields two or even three
angles were insertion is possible.

For turning the valve stem, we finally close the gripper again and rotate the wrench clockwise until we are reaching the joint limits of the UR5 arm 
(which results in approx. $540^\circ$ revolution). 
To address possible failure cases, we repeat the whole insertion procedure until the operator stops the process---which concludes the competition run.
\section{Evaluation}

\begin{figure}
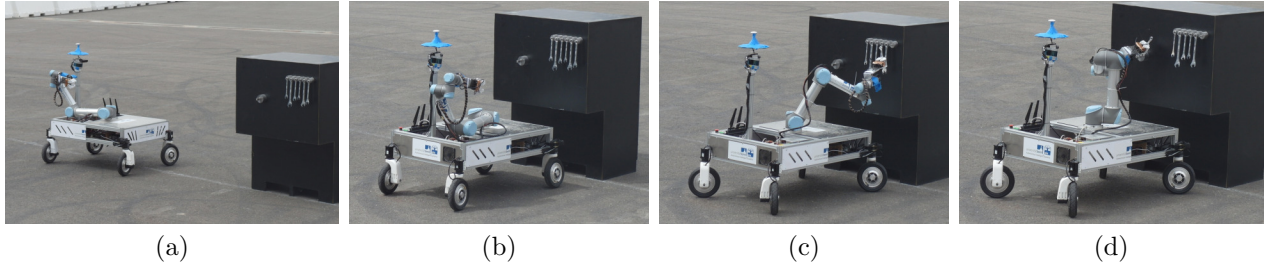

 \centering\begin{maybepreview}%
 \setlength{\tabcolsep}{2pt}%
 \begin{tabular}{cccc}
 \includegraphics[height=2.95cm]{images/run/MBZIRC_NimbRo_Challenge_2_Run_2_2_2017_03_17.jpg} &
 \includegraphics[height=2.95cm]{images/run/MBZIRC_NimbRo_Challenge_2_Run_2_3_2017_03_17.jpg} &
 \includegraphics[height=2.95cm]{images/run/MBZIRC_NimbRo_Challenge_2_Run_2_5_2017_03_17.jpg} &
 \includegraphics[height=2.95cm]{images/run/MBZIRC_NimbRo_Challenge_2_Run_2_6_2017_03_17.jpg} \\
 (a) & (b) & (c) & (d)
 \end{tabular}\end{maybepreview}
 \caption{Our very successful second Challenge~2 run.
 (a) Driving around the panel to the front side; (b) Close approach; (c) Grasping the correct wrench;
 (d) Inserting the wrench and turning the valve.
 }
 \label{fig:eval_run}
\end{figure}

Mario was very successful at the MBZIRC Finals. Our team NimbRo won the ground robot challenge
and later won the Grand Challenge consisting of all three challenges together.
In the following, we analyze the overall performance during the competition and evaluate
the individual components in additional experiments.

\subsection{Performance at the MBZIRC Finals 2017}

During two rehearsal runs, we optimized the panel approach and valve manipulation.
In particular, the practice arena was designed for the other challenges and included
large weights for the net that marked the arena boundary for UAVs. These weights were of
very similar shape and coloring compared to the valve panel. This was not an issue
in the competition runs, since the arena for the ground robot challenge did not have
such a net.

\begin{table}
\centering
\begin{threeparttable}
\caption{System component success rates in the competition runs.}
\label{tab:succ_rate}
  \begin{tabular}{lccccc}
  \toprule
  Competition run	& Locomotion	& Wrench selection\tnote{1}	& Grasping	& Valve detection	& Wrench insertion\\
  \midrule
  Challenge~2, Run~1	& 2/3\tnote{2}	& 4/4				& 2/2		& 1/2\tnote{3}		& 1/1\\
  Challenge~2, Run~2	& 1/1		& 2/2	 			& 1/1		& 1/1 			& 1/1\\
  Grand Ch., Run~1	& 1/1		& 2/2				& 1/1		& 1/1			& 1/1\\
  Grand Ch., Run~2	& 1/1		& 2/2				& 1/1		& 2/2			& 1/2\tnote{4}\\
  \midrule
  Total			& 5/6		&10/10				& 5/5		& 5/6			& 4/5\\
  \bottomrule
  \end{tabular}
  \begin{tablenotes}
  \footnotesize
  \item [1] Two independent pipeline runs in each situation, one per camera.
  \item [2] Missing waypoint (operator mistake).
  \item [3] Wrench was grasped in bad pose (suboptimal motion primitive).
  \item [4] Insertion failed on the first attempt for unknown reason.
  \end{tablenotes}
\end{threeparttable}
\end{table}

\begin{table}
\centering
\begin{threeparttable}
\caption{Top Challenge~2 teams.}
\label{tab:results}
  \begin{tabular}{rlrrc}
  \toprule
  Rank 	& Team 							& Score & Time [sec.]\tnote{1}\\
  \midrule
  1	& University of Bonn 					& 100	& 84\\
  2	& Technical University of Denmark			& 100	& 252\\
  3	& King Abdullah University of Science and Technology	& 45	& 1278\\
  4	& Korea Advanced Institute of Science and Technology	& 35	& 1500\\
  \bottomrule
  \end{tabular}
  \begin{tablenotes}
  \footnotesize
   \item [1] Results provided by MBZIRC organizers during a workshop at ICRA 2017.
  \end{tablenotes}
\end{threeparttable}
\end{table}

We aggregated the information about successes and failures during the competition days in \cref{tab:succ_rate}.
On the first competition day, our run initially suffered from a missing waypoint update
(an operator mistake) and GPS mislocalization.
After calling a reset and resolving both issues, the robot managed to find the panel quickly,
selected and grasped the correct wrench, and started to search for the valve stem.
However, the wrench was not grasped correctly and was not visible in the
pico flexx depth image as expected (see \cref{sec:valve_perception}).
After another reset, Mario performed the entire task successfully with full autonomy.
Including both resets, our time needed was over six minutes.
After the first competition day, this placed us in first place---only DTU-Elektro, Automation and Control Team
from Technical University of Denmark managed to solve the task also in six minutes, closely behind us.

Before our second competition day, we managed to robustify the GPS localization and
improved the valve perception to allow the system to continue even if the wrench
is not correctly localized. In our competition run (see \cref{fig:eval_run}),
Mario solved the entire task
without resets in one minute and 24 seconds,  
which secured the first place for this challenge. A video of this run is available online\footnote{Video: \url{https://youtu.be/TMiTC9wa5S8}}. DTU-Elektro also performed the task autonomously, but 
took over four minutes for their run (see \cref{tab:results}).

For the Grand Challenge, our robots had to perform the ground robot task simultaneously
with the UAV tasks. Again, two attempts were allowed by the rules. Mario solved
the wrench turning task in one minute and 23 seconds on the first attempt, and around five minutes on
the second attempt since we had decided to wait for the landing challenge to complete
before starting Mario. Since the challenges were first ranked separately, this gave us
first rank for the ground robot challenge and a mean rank of 1.7 over all challenges,
yielding the overall first place in the Grand Challenge.

From a more technical standpoint, the navigation based on GPS and LiDAR performed very well.
Our system would have been able to cope with obstacles during global navigation, but this
was not necessary in the flat, open arena.
Panel detection was very reliable once the robot was close enough to have robust measurements
on the black panel surface. Our panel approach trajectory took us quickly around the panel
into the correct approach direction for manipulation.
The wrench detection and selection pipeline had a 100\% success rate in the competition and during on-site
testing---somewhat surprising us, since this was one area where we expected problems due to
wind, lighting conditions, and different wrench models.
Finally, the valve stem perception worked robustly after our first run and allowed
us to quickly insert the wrench and turn the valve.

\subsection{Detailed Timing Analysis}
\label{sec:eval:timing}

\begin{figure}
\centering
\begin{maybepreview}%
\pgfplotstableread{ %
Label Navigate ApproachPanel CyclePanel ArrivePanel SeeWrenches SelectWrench GraspWrench SeeValve DetectValve ContactValve InsertWrench TurnValve SeeValve2 DetectValve2 ContactValve2 InsertWrench2 TurnValve2
Ch4Run2 -6.4 -0.00 -11.2 -13.5 -5.8 -3.4 -11.6 -1.9 -0.7 -5.7 -17.5 -5.0 -15.9 -0.8 -4.4 -16.9 -5.0
Ch4Run1 -0.2 -5.6 -12.9 -13.4 -5.8 -3.3 -11.6 -2.0 -0.8 -5.7 -19.1 -3.0 -00.0 -0.0 -0.0 -00.0 -0.0
Ch2Run2 -7.8 -6.7 -13.0 -13.5 -5.8 -3.3 -11.6 -2.1 -0.8 -5.4 -18.5 -4.0 -00.0 -0.0 -0.0 -00.0 -0.0
Ch2Run1 -7.9 -7.9 -10.7 -14.1 -5.8 -2.5 -11.7 -1.8 -0.7 -5.6 -18.7 -4.0 -00.0 -0.0 -0.0 -00.0 -0.0
}\rundata%
    \begin{tikzpicture}[
      ll/.style={rotate=65, anchor=west,align=left}
    ]

    \begin{axis}[
    width=\linewidth,
    height=7cm,
    xbar stacked,   %
    ytick=\empty,     %
    every node near coord/.style={
      check for zero/.code={
        \pgfkeys{/pgf/fpu=true}
        \pgfmathparse{\pgfplotspointmeta-1.7}
        \pgfmathfloatifflags{\pgfmathresult}{-}{
           \pgfkeys{/tikz/coordinate}
        }{}
        \pgfkeys{/pgf/fpu=false}
      }, check for zero, font=\scriptsize},
    nodes near coords style={font=\tiny, rotate=90, /pgf/number format/.cd, precision=1, fixed zerofill},
    nodes near coords={\pgfmathprintnumber{\pgfplotspointmeta}\,s},
    ytick=data,
    yticklabels from table={\rundata}{Label},
    legend style={at={(axis cs:-65,3.8)},anchor=south},
    legend columns=2,
    transpose legend,
    xlabel={Time until task completion (in seconds)},
    y axis line style={draw opacity=0},
    ytick pos=left,
    xtick pos=left,
    xtick={-130,-120, ..., 0},
    xmin=-130,
    bar width=0.8cm,
    reverse legend,
    y=0.9cm,
    enlarge y limits = 0.15,
    font=\footnotesize,
    ]

    \addplot [fill=orange!20!red, point meta=explicit] table [x=TurnValve2, y expr=\coordindex, meta expr=-\columnx] {\rundata}; 
    \addplot [fill=orange!60!red, point meta=explicit] table [x=InsertWrench2, y expr=\coordindex, meta expr=-\columnx] {\rundata};
    
    \addplot [fill=orange!80!red, point meta=explicit] table [x=ContactValve2, y expr=\coordindex, meta expr=-\columnx] {\rundata};
    \addplot [fill=green!80!black, point meta=explicit] table [x=DetectValve2,y expr=\coordindex, meta expr=-\columnx] {\rundata};
    \addplot [fill=orange!80!yellow, point meta=explicit] table [x=SeeValve2, meta=Label,y expr=\coordindex, meta expr=-\columnx] {\rundata};
    
    \addplot [fill=orange!20!red, point meta=explicit] table [x=TurnValve, meta=Label,y expr=\coordindex, meta expr=-\columnx] {\rundata};\addlegendentry{Turn valve}
    \addplot [fill=orange!60!red, point meta=explicit] table [x=InsertWrench, meta=Label,y expr=\coordindex, meta expr=-\columnx] {\rundata};\addlegendentry{Insert wrench}
    
    \addplot [fill=orange!80!red, point meta=explicit] table [x=ContactValve, meta=Label,y expr=\coordindex, meta expr=-\columnx] {\rundata};\addlegendentry{Contact valve}
    \addplot [fill=green!80!black, point meta=explicit] table [x=DetectValve, meta=Label,y expr=\coordindex, meta expr=-\columnx] {\rundata};\addlegendentry{Detect valve}
    \addplot [fill=orange!80!yellow, point meta=explicit] table [x=SeeValve, meta=Label,y expr=\coordindex, meta expr=-\columnx] {\rundata};\addlegendentry{See valve}
    \addplot [fill=orange!50!yellow, point meta=explicit] table [x=GraspWrench, meta=Label,y expr=\coordindex, meta expr=-\columnx] {\rundata};\addlegendentry{Grasp wrench}
    \addplot [fill=green!40, point meta=explicit] table [x=SelectWrench, meta=Label,y expr=\coordindex, meta expr=-\columnx] {\rundata};\addlegendentry{Select wrench}
    \addplot [fill=orange!25!yellow, point meta=explicit] table [x=SeeWrenches, meta=Label,y expr=\coordindex, meta expr=-\columnx] {\rundata};\addlegendentry{See wrenches}
    \addplot [fill=blue!70, point meta=explicit] table [x=ArrivePanel, meta=Label, y expr=\coordindex, meta expr=-\columnx] {\rundata};\addlegendentry{Arrive panel}
    \addplot [fill=blue!50, point meta=explicit] table [x=CyclePanel, meta=Label, y expr=\coordindex, meta expr=-\columnx] {\rundata};\addlegendentry{Cycle panel}
    \addplot [fill=blue!30, point meta=explicit] table [x=ApproachPanel, meta=Label, y expr=\coordindex, meta expr=-\columnx] {\rundata};\addlegendentry{Approch panel}
    \addplot [fill=blue!15, point meta=explicit] table [x=Navigate, meta=Label, y expr=\coordindex, meta expr=-\columnx] {\rundata};\addlegendentry{Navigate}
    
    \end{axis}
    \end{tikzpicture}\end{maybepreview}
    \caption{System component time measurements during the competition runs.
    Locomotion components are depicted in blue, manipulation components in orange, and perception components in green.
    Only the last successful trial (after two resets) is considered for Challenge~2 Run~1.
    The measurements for Challenge~4 Run~2 include two valve turning attempts, since the first attempt failed.
    }
    \label{fig:timings_run2}
\end{figure}
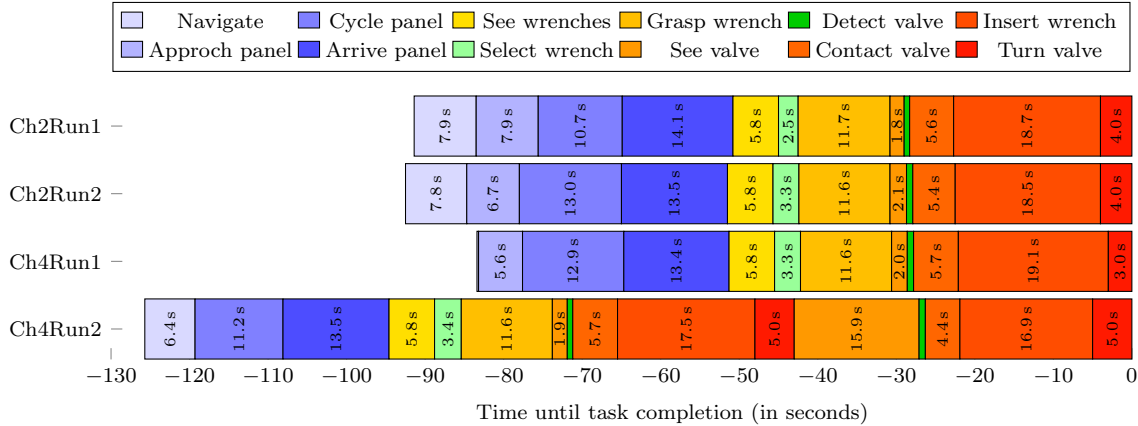

Speed was---as expected prior to the challenge---the criterion determining the winner for Challenge~2, since more than one team solved the task autonomously.
We spent much time accelerating and fine tuning each system component during our tests in Abu Dhabi.
\Cref{fig:timings_run2} shows the 
time needed for each component during our four competition runs. The timings were measured by our high-level state machine.

\label{sec:eval_base}
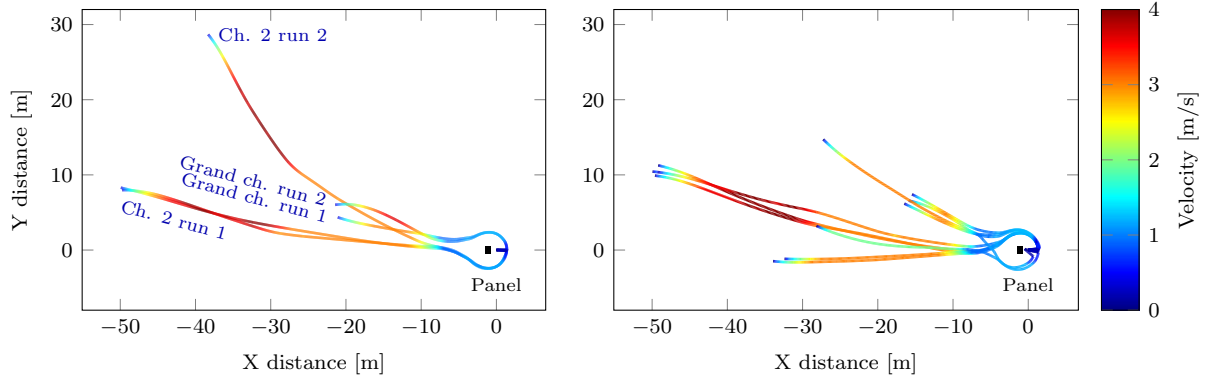
\begin{figure}
 \centering\begin{maybepreview}%
\begin{tikzpicture}[
      font=\footnotesize,
      trajlabel/.style={font=\scriptsize},
]

\begin{axis}[
	x=0.1cm, y=0.1cm,
    ymin=-8, ymax=32,
    height=6cm,
    width=.5\linewidth,
    colormap/jet,
    traj/.style={mesh,point meta=explicit,line width=1pt},
    xlabel={X distance [m]},
    ylabel={Y distance [m]},
    point meta max=4
]

\addplot[traj] table[x=rx,y=ry,meta=velocity] {images/approach/run1_002.txt} node[pos=1.0,anchor=north west,trajlabel,inner sep=0cm,shift={(axis direction cs:0,-1)},rotate=-15] {Ch. 2 run 1};
\addplot[traj] table[x=rx,y=ry,meta=velocity] {images/approach/run1_003.txt};;
\addplot[traj] table[x=rx,y=ry,meta=velocity] {images/approach/run2_2017-03-17-10-45-54_001.txt} node[pos=1.0,anchor=west,trajlabel] {Ch. 2 run 2};;

\addplot[traj] table[x=rx,y=ry,meta=velocity] {images/approach/ch4_run1_2017-03-18-08-14-08_001.txt} node[pos=1.0,anchor=east,trajlabel,rotate=-15,shift={(axis direction cs:0,-0.3)}] {Grand ch. run 1};
\addplot[traj] table[x=rx,y=ry,meta=velocity] {images/approach/ch4_run2_2017-03-18-12-11-46_001.txt} node[pos=1.0,anchor=east,trajlabel,rotate=-15,shift={(axis direction cs:0,0.3)}] {Grand ch. run 2};

\begin{scope}[shift={(axis direction cs:-1.1,0)}]

\fill[black] (axis cs:-0.38,-0.5)  rectangle (axis cs:0.38,0.5);

\end{scope}

\node[trajlabel,anchor=north,shift={(axis direction cs:0,-2.5)}] at (0,0) {Panel};

\end{axis}

\end{tikzpicture}  %
\begin{tikzpicture}[
      trajlabel/.style={font=\scriptsize},
      font=\footnotesize,
]

\begin{axis}[
   	x=0.1cm, y=0.1cm,
    ymin=-8, ymax=32,
    height=6cm,
    width=.4\linewidth,
    colormap/jet,
    colorbar,
    traj/.style={mesh,point meta=explicit,line width=1pt},
    xlabel={X distance [m]},
    colorbar style={ylabel={Velocity [m/s]}},
    point meta max=4
]

\addplot[traj] table[x=rx,y=ry,meta=velocity] {images/approach/testing/arena_test_2017-03-15-11-23-16_001.txt};
\addplot[traj] table[x=rx,y=ry,meta=velocity] {images/approach/testing/arena_test_2017-03-15-11-39-58_001.txt};
\addplot[traj] table[x=rx,y=ry,meta=velocity] {images/approach/testing/arena_test_2017-03-15-11-39-58_005.txt};
\addplot[traj] table[x=rx,y=ry,meta=velocity] {images/approach/testing/demo_2017-03-18-14-09-42_001.txt};
\addplot[traj] table[x=rx,y=ry,meta=velocity] {images/approach/testing/demo_2017-03-18-14-09-42_002.txt};
\addplot[traj] table[x=rx,y=ry,meta=velocity] {images/approach/testing/parkplatz_test_2017-03-16-04-38-21_001.txt};
\addplot[traj] table[x=rx,y=ry,meta=velocity] {images/approach/testing/parkplatz_test_2017-03-16-04-38-21_003.txt};
\addplot[traj] table[x=rx,y=ry,meta=velocity] {images/approach/testing/parkplatz_test_2017-03-16-04-38-21_004.txt};
\addplot[traj] table[x=rx,y=ry,meta=velocity] {images/approach/testing/parkplatz_test_2017-03-16-17-26-35_004.txt};
\addplot[traj] table[x=rx,y=ry,meta=velocity] {images/approach/testing/parkplatz_test_2017-03-16-17-52-01_002.txt};

\begin{scope}[shift={(axis direction cs:-1.1,0)}]

\fill[black] (axis cs:-0.38,-0.5)  rectangle (axis cs:0.38,0.5);

\end{scope}

\node[trajlabel,anchor=north,shift={(axis direction cs:0,-2.5)}] at (0,0) {Panel};

\end{axis}

\end{tikzpicture} %
\end{maybepreview}
 \caption{Panel approach trajectories. Shown are odometry trajectories of different panel approach runs,
 transformed such that the final position in front of the panel is at (0,0).
 Linear velocity of the robot is encoded as color.
 Left: Competition runs. Note that we performed two successful approaches in our first run, with a reset in between.
 Right: On-site practice runs outside of the arena. Best viewed in color.}
 \label{fig:eval:trajectories}
\end{figure}

We made use of our fast omnidirectional base (see \cref{sec:base}) and tuned the first navigation phase to reach speeds of up to 4\,$\frac{m}{s}$.
For close approach and driving around the panel, Mario slows down to avoid control oscillations.
\Cref{fig:eval:trajectories} shows aggregated approach trajectories from our competition and practice runs.
Also the high acceleration of the base (initial acceleration up to $2\,\frac{\textrm{m}}{\textrm{s}^2}$) can be clearly seen.
Indeed, the maximum velocity is close to the limit of $15\,\frac{\textrm{km}}{\textrm{h}}$ ($4.17\,\frac{\textrm{m}}{\textrm{s}}$) given by the rules.
The theoretical minimum time needed to approach the panel given the velocity limit is
$50\,\textrm{m} \left(15 \frac{\textrm{km}}{\textrm{h}} \right)^{-1} = 12\,\textrm{s}$.
Our system reaches the far side of the panel in roughly 27\,s (excluding the slower fine positioning),
which is a very good result considering that our
robot needs to slow down for circling the panel to reach the other side and is limited by acceleration constraints.

Velocity and acceleration of each executed motion during the manipulation subtasks were optimized before the run. Critical motions 
such as grasping the wrench or making contact with the valve and inserting the wrench were executed slower to reduce possible error sources as much as possible.
All other motions were optimized in time to speed up the entire task.
Joint space motions were executed with a speed of up to 2\,$\frac{\textrm{rad}}{\textrm{s}}$, Cartesian space motions
with a speed of up to 1.3\,$\frac{\textrm{m}}{\textrm{s}}$ at the endeffector.

Our perception components needed a relatively short execution time (in total approx. 4\,\% of the task completion time) and were therefore not optimized further.

\subsection{Wrench Perception}

\begin{figure}
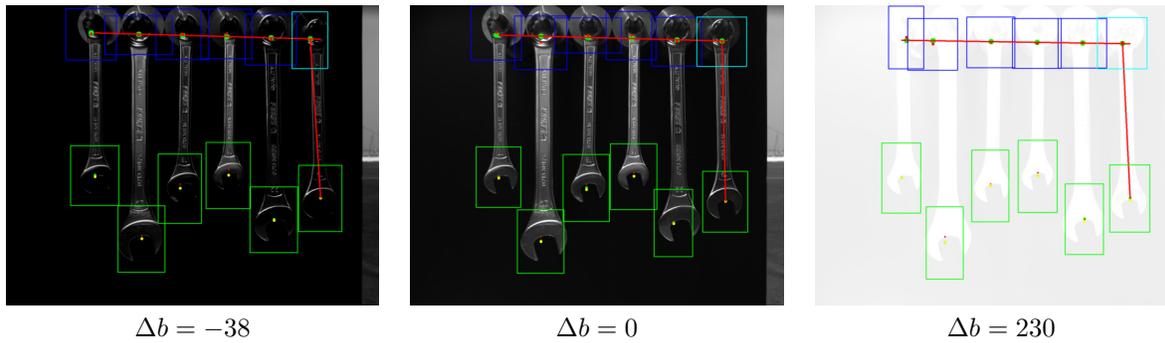

 \centering\begin{maybepreview}
 \begin{tabular}{ccc}
 \includegraphics[angle=-90,trim=0 100 100 450,clip,width=.3\linewidth]{images/wrench_robustness/vis_adapted_-3.png} &
 \includegraphics[angle=-90,trim=0 100 100 450,clip,width=.3\linewidth]{images/wrench_robustness/vis_adapted_0.png} &
 \includegraphics[angle=-90,trim=0 100 100 450,clip,width=.3\linewidth]{images/wrench_robustness/vis_adapted_18.png} \\
 $\Delta b = -38$ & $\Delta b = 0$ & $\Delta b = 230$
 \end{tabular}\end{maybepreview}
 \caption{Example frames of brightness robustness experiment for wrench perception.
 Shown are the orginal image (center), and lower and upper boundary cases with 100\% success rate.}
 \label{fig:eval:wrench_brightness}
\end{figure}

\begin{figure}
 \centering\begin{maybepreview}
 \begin{tikzpicture}[font=\footnotesize]
  \begin{groupplot}[group style={group size=2 by 1},
    height=5cm,width=8.4cm,enlarge x limits=false,
    ymin=0,ymax=1,enlarge y limits=true
    ]
    
    \nextgroupplot[title={Brightness Offset}, xlabel={$\Delta b$},ylabel={Success rate}]
    \addplot+ table[x expr=0.05*255*\thisrow{factor},y expr=(10 - \thisrow{errors})/10] {data/wrench_perception_brightness.txt};
   
    \addplot +[red,thick,mark=none] coordinates {(-38, -1) (-38, 2)};
    \addplot +[black,thick,mark=none] coordinates {(0, -1) (0, 2)};
    \addplot +[red,thick,mark=none] coordinates {(230, -1) (230, 2)};
    
    \nextgroupplot[title={Gaussian Noise}, xlabel={$\sigma$}]
    
    \addplot+ table[x=stddev,y expr=(10 - \thisrow{errors})/10] {data/wrench_perception_noise.txt};
  \end{groupplot}
 \end{tikzpicture}\end{maybepreview}
 \caption{Wrench perception robustness.
 Left: Robustness against brightness changes. Cases shown in \cref{fig:eval:wrench_brightness} are marked with vertical lines.
 Right: Robustness against Gaussian noise with standard deviation $\sigma$ in RGB channels.}
 \label{fig:eval:wrench_robustness}
\end{figure}
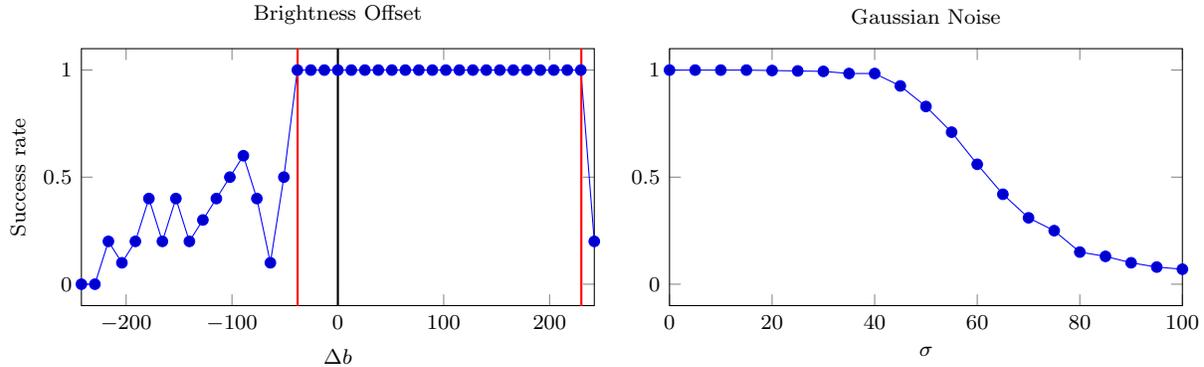

The wrench perception pipeline was evaluated further on the data captured during the MBZIRC competition runs.
Mario performed five wrench grasp pose estimations during the competition, all of which were successful.
Since each estimation was performed using the two PointGrey cameras independently, we obtained ten test images.
We disturbed the images in a controlled way and then measured the success rate over the images.
Success is defined as estimating a grasp point that does not differ by more than 35\,pixels (corresponds to 1\,cm on the panel plane)
from the grasp point estimated during our successful competition runs.
We first disturb the images by adjusting brightness (adding a constant $\Delta b$).
\Cref{fig:eval:wrench_brightness} shows boundary cases, where the method still works,
while \cref{fig:eval:wrench_robustness} reports performance across all $\Delta b$ values.
The method works reliably even with very bright images, while being sensitive to too dark images ($\Delta b < -38$).
Further analysis revealed that the wrench mouth tips are already very dark, so they tend to disappear against the black background.
During the competition, the cameras were set to constant exposure (manually tuned), so a different exposure value could
have resulted in a larger margin of error in both directions.

The method is also fairly robust to Gaussian noise introduced in the grayscale intensity, as it would occur in
low-light and other difficult measurement situations.
\Cref{fig:eval:wrench_robustness} shows a graceful decline in performance for $\sigma > 45$.

\subsection{Valve Registration}

\begin{figure}
 \centering\begin{maybepreview}
 \begin{tikzpicture}[font=\footnotesize]
  \begin{groupplot}[group style={group size=2 by 1},height=5cm,width=8.4cm,enlarge x limits=false]
   
   \nextgroupplot[title={Missing Measurements},xlabel={$p_{\textrm{missing}}$},ylabel={Success rate}]
   \addplot+ table[x=dropout,y=dropsuccess] {data/valve_registration_robustness.txt};
   
   \nextgroupplot[title={Gaussian Noise}, xlabel={$\sigma$}, use units, change x base, x unit=m, x SI prefix=centi]
   \addplot+ table[x=stddev,y=noisesuccess] {data/valve_registration_robustness.txt};
  \end{groupplot}
 \end{tikzpicture}\end{maybepreview}
 \caption{Valve registration robustness.
 Perception success rate is measured on five pico flexx pointclouds recorded during the official competition runs.
 Left: Robustness against artificial missing measurements with drop probability $p_{\textrm{missing}}$.
 Right: Robustness against Gaussian noise on depth measurements.
 }
 \label{fig:valve_registration}
\end{figure}
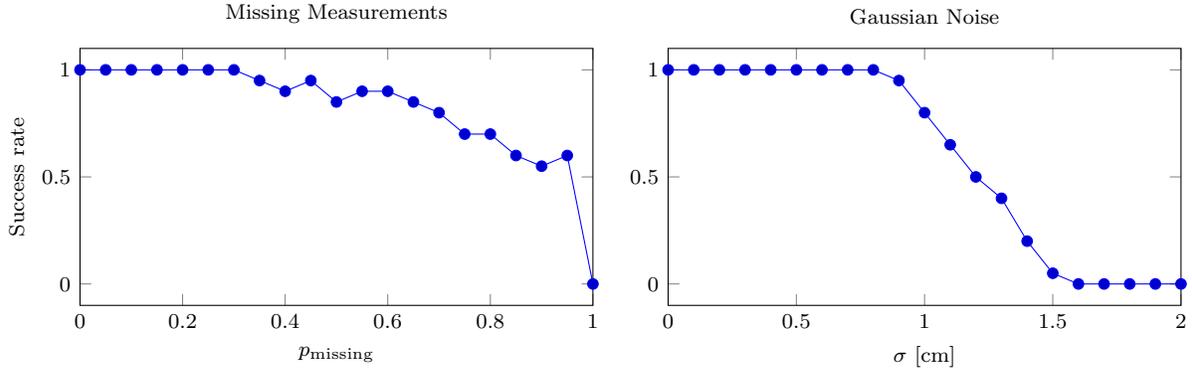

We also evaluated our valve registration component on data from competition runs.
During the four runs, our robot performed four successful valve registrations.
Using these frames as test set, we investigate the robustness of our approach.
In a first experiment, we delete measurements randomly with varying Bernoulli probability $p_{\textrm{missing}}$
and measure the perception success rate across the frames. Here, success is achieved if the valve center
is found and is within $d=1\,\textrm{cm}$ of the prediction from the competition run (which led to successful manipulation).
Each run for a particular value of $p_{\textrm{missing}}$ is repeated five times to reduce variance.
Results can be seen in \cref{fig:valve_registration}.
We can see that the valve registration works robustly with up to half of the possible measurements missing.

In a second experiment, we introduced Gaussian noise on the depth measurements with standard deviation $\sigma$
(see \cref{fig:valve_registration}). The valve registration works robustly up to $\sigma=0.7\,\textrm{cm}$, which
is a good result given that the valve stem measures $1.9\,\textrm{cm}$ in width and $\approx 5\,\textrm{cm}$ in length.

In summary, both wrench perception and valve registration are robust against degradation of the recorded data,
such as they would occur under different lighting conditions or measurement difficulty, and were operated during
the MBZIRC competition with large enough margins to the operating limits.

\section{Conclusion \& Lessons Learned}

Overall, the MBZIRC 2017 was a highly demanding and successful event for us.
In this article, we motivated our design choices, detailed the individual system components,
and reported on our system integration efforts. In addition to system integration, notable
contributions include the far-range panel detection in 3D laser scans, real-time panel pose estimation,
robust controllers for quickly approaching the panel,
the adaption of a state-of-the-art object detection pipeline to the wrench
selection task, parameterized motion primitives for wrench manipulation, and in particular a wrench-insertion method that self-aligns the wrench mouth to the valve stem.

Several of our design decisions turned out to be correct and paid off during
the competition. One such major decision was to focus on execution time.
To this end, we constructed a fast omnidirectional mobile base, which was a major advantage.
This was complemented by the fast Velodyne Puck navigation sensor and the camera-based perception of the wrenches and the valve stem, which took only a fraction of a second. 
The massive computations of the CNN-based object detection were accelerated by an onboard GPU.
We also made significant efforts making sure that the system
did not loose time during manipulation, by tuning each motion to be as fast as
possible. As a side-effect, this also considerably shortened test times
and thus sped up development cycles.

Testing on-site was very important to our success, because factors like wrench motion due to wind and lighting effects were hard to anticipate in advance. We successfully built
a mock-up replica of the panel with wrenches and used it to train the
approach and manipulation phases.

The off-the-shelf UR5 robotic arm has proven itself to be very reliable
and precise. The two teams who successfully completed the task both opted
for the UR5.

We hope that the description of our work will help other researchers to
solve similar problems and thus further advance the state of the art
in field robotics.
 
\subsubsection*{Acknowledgment}

This research was supported by a Challenge Sponsorship from Khalifa University of Science, Technology and Research, Abu Dhabi, United Arab Emirates.

\bibliographystyle{apalike}
\bibliography{paper}

\end{document}